\documentclass[runningheads,a4paper]{llncs}

\usepackage{amssymb}
\usepackage{graphicx}
\usepackage{amsmath}
\usepackage{longtable}

\usepackage{etaremune}

\newcommand{\abs}[1]{\left\lvert #1 \right\rvert}

\usepackage{url}
\urldef{\mailsa}\path|{thounaojam_pg,soujanyo_ug,vivekmeena_ug, debojyoti_rs, anupam}@cse.nits.ac.in|    

\usepackage{tikz}
\usetikzlibrary{shapes.geometric, arrows}

\tikzstyle{startstop} = [rectangle, rounded corners, minimum width=3cm, minimum height=1cm,text centered, draw=black, fill=red!30]

\tikzstyle{io} = [trapezium, trapezium left angle=70, trapezium right angle=110, minimum width=3cm, minimum height=1cm, text centered, draw=black, fill=blue!30]

\tikzstyle{process} = [rectangle, minimum width=9cm, minimum height=1.5cm, text centered, text width=7cm, draw=black, fill=orange!30]
\tikzstyle{decision} = [diamond, minimum width=2cm, minimum height=1cm, text centered, text width=2cm, draw=black, fill=green!30]

\tikzstyle{smallProcess} = [rectangle, minimum width=1cm, minimum height=0.1cm, text centered, text width=3cm, draw=black, fill=orange!30]

\tikzstyle{arrow} = [thick,->,>=stealth]

\begin{document}

\mainmatter  

\title{Introductory Review of Swarm Intelligence Techniques}

\titlerunning{Introductory Review of Swarm Intelligence Techniques}

%
%
%
\authorrunning{Chinglemba et al.}

\author{Thounaojam Chinglemba
\and Soujanyo Biswas
\and Debashish Malakar
\and Vivek Meena
\and Debojyoti Sarkar
\and Anupam Biswas
\thanks{Corresponding author}
}

\institute{Department of Computer Science and Engineering,\\
National Institute of Technology Silchar, Silchar-788010, Assam, India\\
\mailsa\\
}

%
%

\toctitle{Introductory Studies of Swarm Intelligence Techniques}
\tocauthor{Authors' Instructions}
\maketitle

\begin{abstract}
With the rapid upliftment of technology, there has emerged a dire need to ‘fine-tune’ or ‘optimize’ certain processes, software, models or structures, with utmost accuracy and efficiency. Optimization algorithms are preferred over other methods of optimization through experimentation or simulation, for their generic problem-solving abilities and promising efficacy with the least human intervention. In recent times, the inducement of natural phenomena into algorithm design has immensely triggered the efficiency of optimization process for even complex multi-dimensional, non-continuous, non-differentiable and noisy problem search spaces. This chapter deals with the Swarm intelligence (SI) based algorithms or Swarm Optimization Algorithms, which are a subset of the greater Nature Inspired Optimization Algorithms (NIOAs).  Swarm intelligence involves the collective study of individuals and their mutual interactions leading to intelligent behavior of the swarm. The chapter presents various population-based SI algorithms, their fundamental structures along with their mathematical models.

Keywords:  Optimization problems, Optimization Algorithms, Nature Inspired Optimization Algorithm, Swarm Intelligence.
\end{abstract}

\section{Introduction}
    
    Optimization is the simple action to make the best utilization of resources. Mathematically, it is finding the most useful solutions out of all possible solutions for a particular problem. There are two types of optimization problems leaning on whether the variables are \textit{continuous} or \textit{discrete}. Comparatively, continuous optimization problems tend to be easier to solve than discrete optimization. A problem with continuous variables is known as continuous optimization. It aims at achieving maximum efficiency to reduce the cost or increase the overall performance. Variables in continuous optimization are permitted to take on any value within a range of values. In discrete optimization, some or all the variables in a problem are required to belong to a discrete set. Improvements in algorithms coupled with advancements in computing technology have dramatically increased the size and complexity of discrete optimization problems that can be solved efficiently. With the introduction of Swarm Intelligence (SI), conventional methods are less frequently utilized. Still, we need better SI techniques. Optimizing solutions comes with some challenges, for instance, it is time consuming, needs to be updated continuously and can consider only a small range of problems and the solutions obtained may be far from being optimal.
    
    Let us take a small example to visualize the importance of SI over conventional techniques. Suppose, there are four explorers searching for a treasure which is located in a mountain region with several ridges and the treasure is located under one such ridge. The other ridges have small treasures which are not worth the effort. The treasure can be visualized as the optimal solution and the other small regions as the local solutions to the problem. Clearly, the four explorers do not want the small treasures and are definitely going to search for the bigger treasure. They start off with a conventional search for the treasure with a conventional technique known as the gradient descent. They randomly land on a place in the region and go downwards on the nearest slope to get to the nearest ridge and dig it up. What are the chances that they will find the treasure on their first try? It’s obviously quite low. In fact, in the real world, finding optimal solutions with this approach is really time consuming if the data to search in reaches a sufficiently large value. Next, they try a SI approach. One explorer each lands in different locations of the region, and they start exploring locally. But this time, they have phones to converse with each other. They communicate with each other and search areas which are optimally closer to the treasure. Finally, after some 'exploration' they find the actual treasure they are searching for. It is to be noted that since they could converse with each other, they can easily skip areas that have been searched by the others. Also, the area of search increases if we consider the whole unit, and also they are searching locally individually. Simply put, the unit \textit{explores} the regions, and individually, they \textit{exploit} a region to find the treasure. It is a lot quicker and less time consuming than the conventional mechanisms to reach the same solution. Mathematically speaking, optimal solutions refer to the minimums of the curve represented by the problem. There can be problems which have only one global minimum, i.e., there is only one slope and any point on the curve can reach that point if we follow the downward slope of the curve. However, most problems in the real world have multiple minimums, which means, there are local minimums effective to only a particular area in the curve and global minimums which are the optimal solution to the problem (or curve). 
    
    Swarm intelligence can solve complex and challenging problems due to its dynamic properties, wireless communications, device mobility and information exchange. Some examples of swarm are that of bees, ants, birds, etc. They are smarter together than alone and this can be seen in nature too, that creatures when working together as unified system can outperform the majority of individual members when they have to make  a decision and solve problems. Swarm Intelligence is based on clustering individually or adding to existing clustering techniques. Besides the applications to the already present conventional optimization problems, SI is useful in various other problems like communication, medical data-set classification, tracking moving objects, and prediction. To put it simply, SI can be applied to a variety of fields in fundamental research, engineering and sciences. However, to think that an algorithm developed matches the criteria for solving every problem out there in the real world is quite naive. Techniques developed using SI are more problem based, and several changes are made already in present algorithms to reach the solutions of the specific problems. Thus, developing an algorithm without a clear objective (, or problem) is not possible. SI techniques are especially useful to find solutions of non-deterministic polynomial-time problems, which would take exponential time to solve if we take a direct approach to solve the problem. Thus, swarm-based techniques are used to find approximate solutions which are relatively good for the problem.

   Rest of the chapter is organized as follows. Section 2 presents a generic framework for SI techniques and briefed primary components of SI techniques.  Section 3 shows how different SI techniques evolved with time. Section 4 briefly explains working principles of prominent SI techniques. Section 5 highlights various application domains of SI techniques. Lastly, section 6 concludes highlighting different nuances, shortcomings, and applicability of SI techniques. 
    
  \section{Generic Framework of Swarm Intelligence Techniques}
\begin{figure}[t]
      \centering
          \begin{center}
        \begin{tikzpicture}[node distance=2.2cm]
            \node (start) [startstop] {Start};
            
            \node (p1) [process, below of = start] {Apply Conventional Techniques like Random Walk, Grid search to generate solution};
            
            \node (p2) [process, below of = p1] {Compute Objective Function};
            
            \node (p3) [decision, below of = p2, yshift=-1cm] {Check for Convergence};
            
            \node (p4) [process, below of = p3, yshift=-1cm] {Optimal Solution};
            
            \draw [arrow] (start) -- (p1);
            \draw [arrow] (p1) -- (p2);
            \draw [arrow] (p2) -- (p3);
            \draw [arrow] (p3) -- node[anchor=west] {yes} (p4);
            \draw [arrow] (p3) -- node[anchor=south] {no} +(6,0) |- (p1);

        \end{tikzpicture}
    \end{center}
      \caption{Generic Flowchart of optimization using Traditional Techniques}
      \label{fig:my_label1}
  \end{figure}
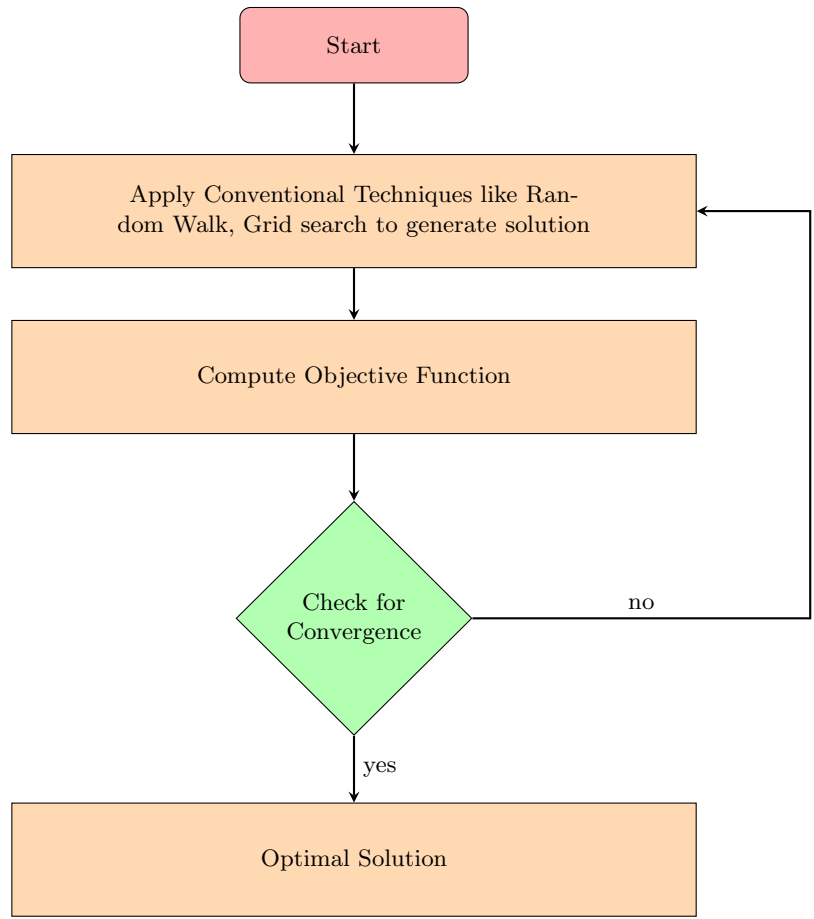

  When we need to find a solution to an optimization problem, we first observe the attributes, constraints, whether it will have a single objective or multiple objectives etc. With traditional optimization, we try to find the solution by traditional methods such as brute force, simulation or apply conventional optimization techniques which is time consuming, requires huge computation cost and requires frequent human intervention. To understand how we approach an optimization problem conventionally, a flow chart is  shown in Fig.~\ref{fig:my_label1}.

    As conventional techniques work best for finding solutions to simple continuous, linear optimization problems, real world optimization are often non-linear and discontinuous in nature. When solving a real world optimization problem complication such as a large number of local solution and constraints, discrete variables, multiple objectives, deceptive search space etc. needs to be addressed. When we plug in these conventional techniques in the real world problem it leads to problems such as non-optimal solutions, being stuck on a local solution, low convergence rate etc. A need to develop an intelligent system to efficiently solve these real world problems have become a necessity. To develop these intelligent systems over the years, researchers and scholars have taken inspiration from various phenomena of nature as these phenomena always tend to self-organize even in the most complex environments of nature. Swarm Intelligence techniques are a subset of these Nature Based algorithms which is based on the swarming behavior of insects, birds, fishes, animals etc. to develop a collective intelligence, self-organized way of solving complex, discontinuous, nonlinear systems efficiently with higher convergence and less time consumption as compared to conventional techniques. In Swarm Based algorithms members of the swarm interact with each other and the environment to develop a collective intelligence.

        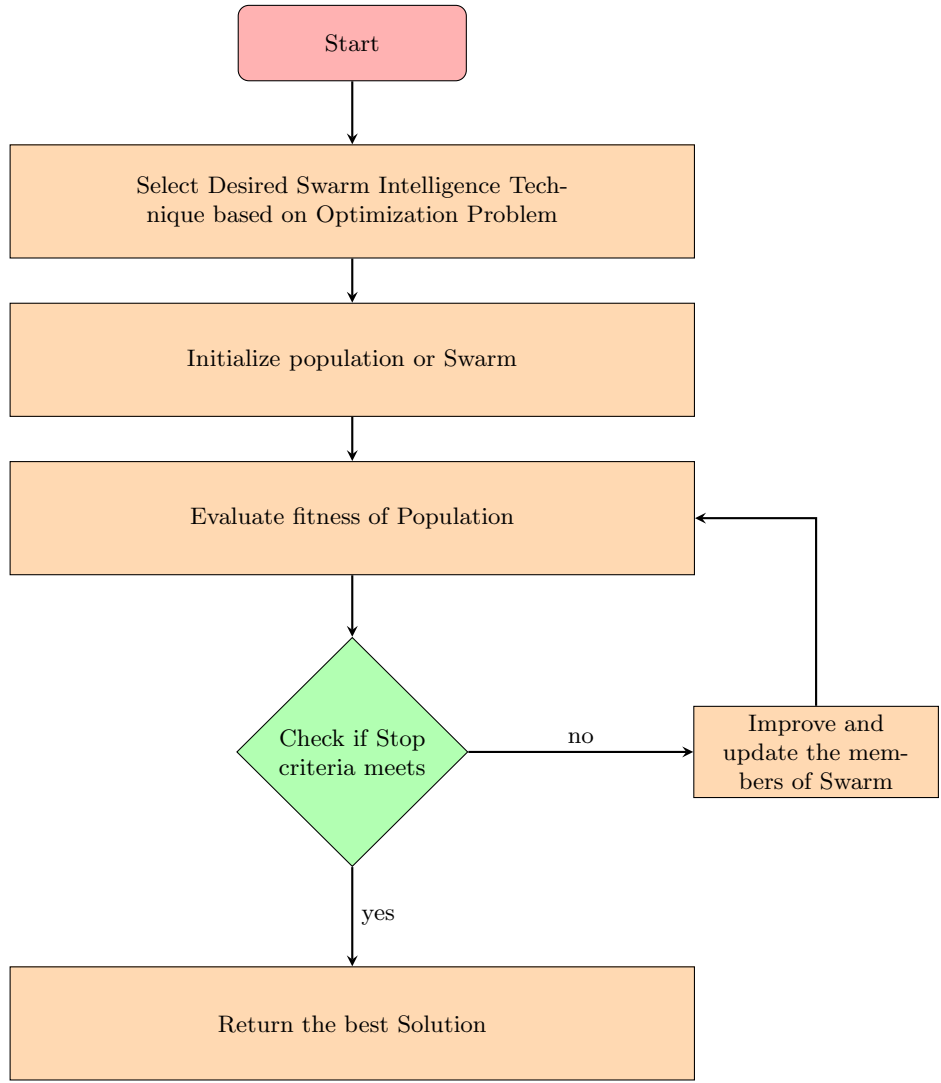
\begin{figure}[!ht]
        \centering
        \begin{center}
        \begin{tikzpicture}[node distance=2.1cm]
            \node (start) [startstop] {Start};
            
            \node (p1) [process, below of=start] {Select Desired Swarm Intelligence Technique based on Optimization Problem};
            
            \node (p2) [process, below of=p1] {Initialize population or Swarm};
            
            \node (p3) [process, below of=p2] {Evaluate fitness of Population};
            
            \node (d1) [decision, below of=p3, yshift=-1cm] {Check if Stop criteria meets}; 
            
            \node (p4) [process, below of=d1, yshift=-1.5cm] {Return the best Solution};
            
            \node (p5) [smallProcess, right of=d1, xshift=4cm] {Improve and update the members of Swarm};
            
            \draw[arrow] (start) -- (p1);
            \draw[arrow] (p1) -- (p2);
            \draw[arrow] (p2) -- (p3);
            \draw[arrow] (p3) -- (d1);
            \draw[arrow] (d1) -- node[anchor=west] {yes} (p4);
            \draw[arrow] (d1) -- node[anchor=south] {no} (p5);
            \draw[arrow] (p5) |- (p3);
        \end{tikzpicture}
    \end{center}
        \caption{Generic Flowchart of Swarm Optimization Techniques}
        \label{fig:my_label2}
    \end{figure}
    
    Over the years as more and more Swarm Intelligence techniques are developed and used, let us look at some important aspects of these swarm based algorithms:
    
    \begin{itemize}
    \item \textbf{Exploration}: One of the most important aspects of every swarm intelligence techniques is its exploration also known as diversification. Exploration traverses the search landscape to search for new solutions which are different from the current solution. This helps us in finding solution which is better than the current solution and helps us in diversifying our solution. It also helps us to escape from a local solution to find new and better solutions. But we should be careful as too much exploration will lead to slow convergence, which is not desired at any swarm intelligence techniques.
    \item \textbf{Exploitation}: While exploration traverses the whole landscape, exploitation, also known as intensity, focuses on a local search area, which will lead to finding better solution in the local search space. This will lead to high convergence in our algorithms but too much exploitation will lead to being stuck in a local solution only. So, a fine balance is needed between the exploration and exploitation~\cite{blum2003metaheuristics} of every swarm intelligence techniques and based on the optimization problem we can fine tune the exploration and exploitation to our requirement.
    \item \textbf{Convergence}: Another important aspect of swarm based algorithms is its high convergence rate. Convergence rate can be defined as the speed in which our algorithm converges to a solution over some iterations beyond which a repeating sequence is generated. While some algorithms use defined global best to converge faster~\cite{kennedy1995particle} to a solution while other algorithms use their own exploration and exploitation~\cite{yang2010firefly} to converge to a solution.
    \item \textbf{Randomization}: In most swarm intelligence techniques, randomization parameters are used for better exploration in the search landscape to find alternate solutions which might be better than the current solutions.
    \end{itemize}

    As we try to understand these swarm based algorithms and how they work collectively as a member of a swarm to achieve collective intelligence, we look at a generic framework of how these swarm based techniques operates  in Fig.~\ref{fig:my_label2}.

\section{Evolution of Swarm Intelligence Techniques}
    
    As swarm based optimization techniques provide self-organized, collective, intelligent and faster convergence in solving complex, discontinuous, nonlinear systems, researchers and scholar are looking to develop novel swarm based optimization techniques that are inspired by biological~\cite{binitha2012survey}, physical~\cite{biswas2013physics} and chemical phenomena~\cite{houssein2019nature} to solve various optimization problems efficiently and effectively compared to traditional techniques. As of now there are more than 140 optimization techniques based on natural phenomenon to solve various optimization problems in field of Science, Medicine, Artificial Intelligence, Engineering etc. Some of the popular swarm intelligence algorithms that have been developed over the years are listed in Table 1.
    
\begin{longtable}{|p{0.1\linewidth}|p{0.4\linewidth}|p{0.4\linewidth}|}
\caption{Some of the popular bio-inspired meta heuristic algorithm inspired by swarm intelligence}
\\ \hline 
\textbf{Year} & \textbf{Algorithm proposed} & \textbf{Inspiration} \\ \hline 
\endfirsthead

\multicolumn{3}{c}%
{} \\
\hline \multicolumn{1}{|c|}{\textbf{Year}} & \multicolumn{1}{c|}{\textbf{Algorithm Proposed}} & \multicolumn{1}{c|}{\textbf{Inspiration}} \\ \hline 
\endhead

\hline \multicolumn{3}{|r|}{{Continued on next page...}} \\ \hline
\endfoot

\endlastfoot

2021 & Flamingo Search Algorithm~\cite{zhiheng2021flamingo} & It is based on migratory and foraging behaviour of flamingos \\ \hline
2021 & Horse herd Optimization Algorithm~\cite{miarnaeimi2021horse} & It implements what horses do at different ages using six important features: grazing, hierarchy, sociability, imitation, defense mechanism and roam. \\ \hline
2020 & Chimp Optimization Algorithm~\cite{khishe2020chimp} & It is inspired from the individual intelligence and sexual behaviours of chimps, when they find a group to be in. \\ \hline
2020 & Black Widow Optimization Algorithm~\cite{hayyolalam2020black} & It is based especially on the interesting sibling cannibalism behaviour offered by Black Widow spiders. \\ \hline
2020 & Sparrow Search Algorithm~\cite{xue2020novel} & This algorithm is based on intelligent techniques used by sparrows to search for food depending on the situation they are in. \\ \hline
2020 & Rat Swarm Optimizer~\cite{dhiman2020novel} & It is inspired by the chasing and attacking behaviour of rats \\ \hline
2019 & The Sailfish Optimizer~\cite{shadravan2019sailfish} & This algorithm uses sailfish population while searching, and sardines population for diversifying the search space, based on a group of hunting sailfish. \\ \hline
2018 & Meerkat Clan Algorithm~\cite{al2018meerkat} & It is based on Meerkat with their exceptional intelligence, tactical organizational skills, and remarkable directional cleverness in its traversal of the desert when searching for food \\ \hline
2018 & Grasshopper Optimization Algorithm~\cite{mirjalili2018grasshopper} & It is inspired by the foraging and swarming behaviour of grasshoppers. \\ \hline
2017 & Salp Swarm Algorithm~\cite{mirjalili2017salp} & It is inspired by the swarming behaviour of salps in oceans. \\ \hline
2017 & Camel Herds Algorithm~\cite{al2017camel} & This algorithm is based on camels, and how they have a leader for each herd and how they search for food and water depending on humidity of neighbouring places. \\ \hline
2017 & Duck Pack Algorithm~\cite{wang2017duck} & It is based on the foraging behaviours of ducks depending on imprinting behaviour and food orientation. \\ \hline
2016 & Dragonfly Algorithm~\cite{mirjalili2016dragonfly} & It is based on the static and dynamic behaviour of dragon flies. \\ \hline
2016 & Sperm Whale Algorithm~\cite{ebrahimi2016sperm} & It is based on the sperm whale's lifestyle. \\ \hline
2016 & Dolphin Swarm algorithm~\cite{wu2016dolphin} & It is based on the biological characteristics and living habits such as echolocation, information exchanges, cooperation, and division of labor of Dolphins. \\ \hline
2016 & Crow Search Algorithm~\cite{askarzadeh2016novel} & It is based on how crows search for food, and hide their food from other crows and remember their hiding places. \\ \hline
2015 & Ant Lion Optimizer~\cite{mirjalili2015ant} & This algorithm mimics the hunting nature of ant-lions in nature. \\ \hline
2015 & Elephant Herding Optimization~\cite{wang2015elephant} & It is based on the herding behaviour of elephants, different group elephants living under a matriarch. \\ \hline
2015 & Moth-flame Optimization algorithm~\cite{mirjalili2015moth} & It is based on the navigation method of moths in nature called transverse orientation. \\ \hline
2014 & Grey Wolf Optimizer~\cite{mirjalili2014grey} & It mimics the living hierarchy and hunting behaviour of grey wolves in nature. \\ \hline
2014 & Pigeon Optimization algorithm~\cite{goel2014pigeon} & It is based on the swarming behaviour of passenger pigeons. \\ \hline
2014 & Spider Monkey Optimization Algorithm~\cite{bansal2014spider} & It is inspired by the Fission-Fusion social structure of spider monkeys during foraging. \\ \hline
2013 & Spider Optimization~\cite{cuevas2013swarm} & It is based on the cooperative characteristics of social spiders. \\ \hline
2012 & Bacterial Colony Optimization~\cite{niu2012bacterial} & It is based on the life cycle of a bacteria named, E. Coli. \\ \hline
2012 & Zombie Survival Optimization~\cite{nguyen2012zombie} & It is based on the foraging behaviour of zombies, when they find a hypothetical airborne antidote which cures their ailments. \\ \hline
2010 & Bat Algorithm~\cite{yang2010new} & It is based on the echolocation of micro-bats in nature, with varying pulse rates of emission and loudness. \\ \hline
2010 & Termite Colony Optimization & It is based on the intelligent behaviour of termites. \\ \hline
2010 & Fireworks Algorithm~\cite{tan2010fireworks} & In this method, two types of explosion processes are performed, and the diversity of them are kept based on fireworks. \\ \hline
2009 & Cuckoo Search~\cite{yang2009cuckoo} & It is inspired by the behaviour of cuckoos to lay eggs in the nests of birds of other species. \\ \hline
2009 & Gravitational Search Algorithm~\cite{rashedi2009gsa} & It is based on Newton's laws and laws of gravity to search for solutions. \\ \hline
2009 & Glowworm Swarm Optimization~\cite{krishnanand2009glowworm} & It simulates the behavior of lighting worms or glow worms. \\ \hline
2008 & Fast Bacterial Swarming Algorithm~\cite{chu2008fast} & This algorithm combines the foraging behaviour of E. Coli from Bacteria Colony Algorithm and flocking mechanism of birds from Particle Swarm Algorithm. \\ \hline
2007 & Firefly Algorithm~\cite{yang2010firefly} & It is inspired from the fireflies in nature. \\ \hline
2006 & Cat Swarm Optimization~\cite{chu2006cat} & It is based on the behaviour of cats, and includes two sub-processes -- tracing mode and seeking mode. \\ \hline
2005 & Artificial Bee Colony~\cite{hancer2012artificial} & This algorithm simulates how honey bee colonies work, including the employed, workers and scouts. \\ \hline
2004 & Honey Bee Algorithm~\cite{wedde2004beehive} & It is based on how honey bees forage for food in nature. \\ \hline
1995 & Particle Swarm Optimization~\cite{kennedy1995particle} & It is based on how flocks of birds move in the world and search for food, both individually and considering the entire group. \\ \hline
1992 & Ant Colony Optimization~\cite{dorigo2006ant} & This algorithm is based on the pheromone based searching done by ants in nature and how they keep track of the food found using their pheromones as location trackers. \\ \hline

\end{longtable}

    As we can see from Table 1, there are various novel nature based optimization techniques to choose from. While some algorithms focus on exploration and exploitation, some focus on faster convergence and parameter tuning etc,. So, before choosing an optimization technique, we need to understand what type of optimization problem we are dealing with, the constraints associated with the optimization problem, the attributes and the mechanism of the algorithm. 
    
    As the nature of optimization problem is different with one another, it has become a necessity to either modify an existing SI algorithm, hybridize with another algorithm or to develop a novel SI based algorithm to obtain the best optimal solution of different optimization problem as it is impossible for a single algorithm to get the best solution for every optimization problem as the nature of the problems are complex in nature.
    
    \section{\Large{Prominent Swarm Intelligence Algorithms}}
    Swarm intelligence techniques in comparison with traditional algorithms provide a novel way to address complex problems and processes efficiently and effectively. These algorithms are based on the swarming behaviours of various insects, birds, animals found in nature where they work collectively to achieve a collective intelligence. While some algorithms work towards  achieving faster convergence, some focuses on the exploration and exploitation of optimization problem. In this section, we will briefly discuss some swarm intelligence algorithms and analyze them.
    
    \subsection{Particle Swarm Optimization}
    Particle Swarm Optimization was introduced by Kennedy and Eberthart in 1995~\cite{kennedy1995particle}, by observing the social foraging behavior of animals such as flock of birds or school of fishes. In flocking of birds, whether it is flocking towards a destination or searching for food, the goal is achieved by cooperating with all the birds in the flock. Each bird learns from its current experience and updates it with the other birds to achieve its goal. The birds follow a leader which is closer to the best solution and any changes made by the leader is followed by all the birds in the flock hence achieving a collective, intelligent and a self-organized behavior as a swarm. Observing this, Kennedy and Eberthart summarized the Particle Swarm Optimization into two equations, one for the position of every particle in the swarm and other equations for the velocity of the particles in the swarm.
    
    In Particle Swarm Optimization we initialize a group of particle randomly and then search for the best optimal solution over many iterations. Each particle in a swarm maintains three things its personal best solution also known as pbest, its global best solution also known as gbest and its current direction. To search for the best optimal solution each particle in the swarm needs to update the velocity equation and the position equation over many iterations. The velocity and position  of a particle, $V_{i}$ and $X_{i}$, respectively can be updated over the iterations as follows:
    \begin{equation}
        V_{i}(t+1)=V_{i}(t)+C_{1} R_{1}\left[X_{p b}(t)-X_{i}(t)\right]+C_{2} R_{2}\left[X_{g b}(t)-X_{i}(t)\right] 
    \end{equation}
    \begin{equation}
        X_{i}(t+1)=X_{i}(t)+V_{i}(t+1)
    \end{equation}
    
    where $R_{1}$, $R_{2}$ are two random vector in the range (0,1) and $C_{1}$, $C_{2}$ are cognitive acceleration coefficient and social acceleration coefficient respectively.\\
    In equation (1), $V_{i}(t)$ denotes the current direction of the particle, $X_{p b}(t)-X_{i}(t)$ describes the cognitive component of the particle where, $X_{p b}(t)$ denotes the position vector of a particle pbest and $X_{g b}(t)-X_{i}(t)$ describes the social component of the particle where, $X_{g b}(t)$ denotes the particle gbest. The new velocity and position of a particle, $V_{i}(t+1)$ and $X_{i}(t+1)$ respectively can be calculated over many iterations to get our desired solution.
    
    Over the years researchers have been proposing new methods to the PSO algorithm due to the simplicity of the algorithm, its low computation cost and effectiveness as compared with traditional optimization algorithms, many algorithms has been proposed which improve the PSO algorithm further. Some of the proposed modification are in the form of hybridization with other nature based optimization algorithms like Genetic Algorithm, ACO~\cite{sait2013cell,kuo2013integration,chen2011solving} etc., modification of the PSO algorithms like fuzzy PSO, QPSO~\cite{jau2013modified} etc. and even extension of PSO algorithms to fields like discrete~\cite{chen2012discrete} and binary optimization. Some of the modification done to the PSO algorithm over the years are given in table 2.
    

\begin{longtable}{|p{0.1\linewidth}|p{0.35\linewidth}|p{0.35\linewidth}|}
\caption{Some modified algorithms based on the PSO algorithm}
\label{tab:my_label}
\\ \hline
\hline \multicolumn{1}{|c|}{\textbf{Year Published}} & \multicolumn{1}{c|}{\textbf{Algorithm Name}} & \multicolumn{1}{c|}{\textbf{References}} \\ \hline 
\endfirsthead

\multicolumn{3}{c}%
{} \\
\hline \multicolumn{1}{|c|}{\textbf{Year Published}} & \multicolumn{1}{c|}{\textbf{Algorithm Name}} & \multicolumn{1}{c|}{\textbf{References}} \\ \hline 
\endhead

\hline \multicolumn{3}{|r|}{{Continued on next page...}} \\ \hline
\endfoot

\endlastfoot

2018 & PSO-CATV & Time-varying Cognitive avoidance PSO~\cite{biswas2018particle}\\ \hline
2013 & PSOCA & Cognitive avoidance PSO~\cite{biswas2013particle}\\ \hline
2013 & MOPSO & Multiple objective PSO~\cite{qiu2013novel} \\ \hline
2013 & PSO-RANDIW  & Random weighted PSO~\cite{biswas2013improved} \\ \hline
2013 & PSO-GA  & Hybridization of PSO with GA~\cite{kuo2013integration}\\ \hline
2013 & PSO-SA  & Hybridization of PSO with SA~\cite{chen2011solving}\\ \hline
2013 & QPSO  & Quantum-behaved PSO~\cite{jau2013modified} \\ \hline
2012 & DPSO & Discrete PSO~\cite{chen2012discrete} \\ \hline
2011 & PSO-ACO & Hybridization of PSO with ACO~\cite{sait2013cell} \\ \hline
2011 & CPSO & Chaotic PSO~\cite{chuang2011chaotic} \\ \hline
\end{longtable}
    
    \subsection{Firefly Algorithm}
    Inspired by the flashing pattern and behavior of fireflies at night, Yang proposed the Firefly algorithm (FA)~\cite{yang2010firefly} in 2008. The flashing light in fireflies serves two purposes; one is to attract the mating partners and the other is to warn of predators. Yang formalized the Firefly algorithm based on the flashing lights and how each firefly is attracted to one another based on the flashing lights. To formalize the FA algorithm Yang idealized some characteristics of the fireflies as:
    \begin{enumerate}
    \item 	The fireflies are considered as unisex i.e. fireflies are only attracted to one another based on the flashing pattern not on the gender of the firefly.
    \item 	The attractiveness of the firefly is directly proportional to the light intensity i.e. a firefly will move towards a brighter one and if there is no light intensity, it will perform a random walk.

    \item   The light Intensity of the fireflies are determined from the objective function.
\end{enumerate}
  With these assumptions, Yang proposed the updated position of a firefly say $i$ to a brighter firefly $j$ at iteration $t+1$ is given as:
  \begin{equation}
   x_{i}^{t+1}=x_{i}^{t}+\beta_{0} e^{-\gamma r_{i j}^{2}}\left(x_{j}^{t}-x_{i}^{t}\right)+\alpha \epsilon_{i}^{t},
  \end{equation}
  
  In the above equation, $x_{i}^{t}$ represents the initial position of the firefly $i$, $\beta_{0} e^{-\gamma r_{i j}^{2}}\left(x_{j}^{t}-x_{i}^{t}\right)$ represents the attractiveness of the firefly $x_{i}$ to another firefly $x_{j}$, $\beta_{0}$ represents the brightness at r=0 and $\alpha \epsilon_{i}^{t}$ represents the randomization term with $\alpha$ representing a randomization parameter in which $\epsilon_{i}^{t}$ is a vector of random numbers derived from a Gaussian, uniform or others distribution.
  
  As the FA algorithm is based on the attraction and attractiveness decreases with the distance between fireflies, the whole population can be divided into sub groups which in turns provides an efficiently and faster way of searching for the global best and local best in the search landscape. Other important aspect of FA algorithm is that can be reduced to other meta-heuristic algorithms  such as SA, DE, APSO thereby having the characteristics of all these algorithms. Over the years FA algorithm has either been modified or hybridize to obtain better results, low computation cost etc. A detailed survey on firefly algorithm and its variants are done by Fister et al.~\cite{fister2013comprehensive}.

    \subsection{Bacteria Colony Algorithm}
    The Bacteria Colony Optimization is based on how bacteria move about in the environment. The major inspiration comes from a certain bacteria called \emph{E. Coli}. This method of optimization follows how bacteria move with their flagellum, how only the ones better at searching survive and with the change of environment, the weaker ones die, and how new offsprings are made from the stronger ones and the cycle continues as bacteria search for more optimal conditions to survive longer.

    There are four major characteristics of bacteria which make up the bacteria colony optimization.
    \begin{itemize}
        \item \textbf{Chemotaxis:} This model very uniquely reproduces how bacteria, especially \emph{E. Coli} move in the environment by the movement of their flagellum, or whip-like structures present in the bacteria. Thus, a better individual means that it has better chances to survive and reproduce.
        \item \textbf{Elimination, reproduction and migration:} We said that natural selection offers the survival of only the bacteria that can search better for the nutrition around. The weaker ones will be eliminated, and the stronger ones reproduce to give better offspring and this cycle will repeat. Chemotaxis, elimination and reproduction is followed by a process called migration. After the nutrition of a place has depleted, the bacteria have to migrate to a newer place with abundant nutrition to continue the former three processes.
        \item \textbf{Communication:} In the bacteria colony model, we find individuals communicating with its neighbours, random individuals in the group, and also with the whole group. This makes the search for better nutrition more efficient.
    \end{itemize}

    The bacteria optimization model can be formulated combining three sub-models:
    \begin{itemize}
        \item In the Chemotaxis and Communication model, bacteria run and tumble and communicate with each other. The position of the bacteria can be formulated using the formula:
        \begin{equation}
            Position_{i}(T) = Position_{i}(T - 1) + R_{i} * (Ru_{Infor}) + R \Delta (i),
        \end{equation}
        \begin{equation}
            Position_{i}(T) = Position_{i}(T - 1) + R_{i} * (Tumb_{Infor}) + R \Delta (i)
        \end{equation}

        \item The Elimination checks that only the bacteria that has an energy level greater than a given energy level can survive. Mathematically, it can be determined as:
        \begin{equation}
            if \hspace{2pt} L_{i} > L_{given}, and i \in healthy, then \hspace{2pt} i \in Candidate_{repr},
        \end{equation} 
        \begin{equation}
            if \hspace{2pt} L_{i} < L_{given}, and i \in healthy, then \hspace{2pt} i \in Candidate_{eli},
        \end{equation}
        \begin{equation}
            if \hspace{2pt} i \in unhealthy, then i \in Candidate_{eli}.
        \end{equation}

        \item Migration model tells the bacteria to move to a newer location with better nutrition, using the formula:
        \begin{equation}
            Position_{i}(T) = rand * (ub - lb) + lb
        \end{equation}
    \end{itemize}
    
    \subsection{Crow Search Algorithm}
    With recent advances in the field of swarm intelligence, the Crow search algorithm, as its name suggests, is a novel method which mimics how crows act in the environment. It is majorly inspired from the fact that crows are intelligent and how they are very efficient in hiding their food in different places. They can remember the location of the food they have hidden for months. Also, interestingly, it has also been observed that crows also keep track of the hiding places of other crows so that they can steal food. To counter this, whenever a crow finds that another crow is following it, they fly away far from the hiding place to trick the other crow and save its food. The crow search algorithm, very cleverly mimics these behaviours of crows to create a very efficient algorithm.

    While implementing this algorithm, two matrices, one for the crows, and the other for the memory of the hiding places of the crows are kept. These keep track of the location of the crow and also the location of the hiding place in the search space. 

    \begin{center}
        \[
        \begin{bmatrix}
            x_{11}^{t} & x_{12}^{t} & \ldots & x_{1d}^{t}\\
            x_{21}^{t} & x_{22}^{t} & \ldots & x_{2d}^{t}\\
            \vdots & \vdots & \vdots & \vdots\\
            x_{N1}^{t} & x_{N2}^{t} & \ldots & x_{Nd}^{t}
        \end{bmatrix}
        \]
    \end{center}

    During the movement of crow \emph{i} towards crow \emph{j}, two things can happen - 
    If crow \emph{j} is unable to discover that crow \emph{i} was following it, then crow \emph{i} updates its position as follows: 
    \begin{equation}
        x_{i}(t+1) = x_{i}(t) + r_{i} * fl_{i}(t) * (m_{j}(t) - x_{i}(t))
    \end{equation}
    
    Where \(r_{i}\) is a random number, and \emph{fl} decides if the search will be local or global. If $fl > 1$ then the crow \emph{i} moves far away from crow \emph{j}, and vice versa.
    If crow \emph{j} is able to detect crow \emph{i} following it, then it moves away from its hiding place. The new position of crow \emph{j} is now a random position on the search space.

    This whole equation can be summarised into two parts based on a value which is called \emph{Awareness Probability}. If it is high, then the search happens to be global, and if it is lowered then the search is local. This happens because if a crow can search for another crow better then it can track long distances better and search for more optimal food.

    \subsection{Grey wolf optimization}
    
    Grey wolf optimizer (GWO) ~\cite{mirjalili2014grey} was proposed by Mirjali in 2014 after taking inspiration from hunting patterns of grey wolves in nature. He also observed the hierarchy of leadership among grey wolves in a pack. Grey wolves live and hunt for prey together in a group. First, the pack tracks the prey and chases it, after which it encircles the prey and starts attacking until the prey stops moving. Taking inspiration from this he proposed the GWO which offers high convergence speed, simple and greater precision as compared to other nature based optimization algorithms. 
    In GWO Mirjali categorize the wolves in the pack into four groups namely the alpha ( ), beta ( ), delta ( ), and lastly the omega ( ). The alpha ( ) is considered as the individual which has the highest authority followed by the beta ( ) and the delta ( ). The rest of the grey wolves are collectively considered as the omega ( ) of the pack.
    
    \begin{itemize}
        \item \textbf{Encircling prey:} In GWO the grey wolves first track and surrounds the prey. This is the encircling part of the algorithm. The mathematical equations representing this is given below:
    \begin{equation}
        D = \abs{\Vec{C}.\Vec{X}_{p}(t)-\Vec{X}(t)}
    \end{equation}

    \begin{equation}
        \Vec{X}(t+1)=\Vec{X}_{p}(t) - \Vec{A}.\Vec{D}
    \end{equation}

    where $\Vec{A}$ and $\Vec{C}$ are coefficient vectors, $t$ denotes the current iteration, $\Vec{X}_{p}$ is the position vector of prey, and $\Vec{X}$ denotes the position vector of a grey wolf.

    The vectors $A$ and $C$ are calculated as follows:
    \begin{equation}
        \Vec{A} = 2\Vec{a}.\Vec{r_1} - \Vec{a}    
    \end{equation}
    
    \begin{equation}
        \Vec{C} = 2.\Vec{r_2}    
    \end{equation}

    where components of $\Vec{a}$ are linearly decreased from $2$ to $0$ over the course of iterations and $\Vec{r_1}$ and $\Vec{r_2}$ are random vectors in $[0,1]$. \\
    
        \item \textbf{Hunting:} After the encircling phase is done, the hunting phase of the algorithm is started. The alpha of the pack leads the pack in hunting for prey followed by the beta and the delta. In GWO to simulate these conditions, it assumes that the alpha has better knowledge about the prey location. The mathematical equations regarding the hunting phase are given below:
        
        \begin{equation}
            \Vec{D_{\alpha}}=\abs{\Vec{C_1}.\Vec{X_{\alpha}} - \Vec{X}}, 
            \Vec{D_{\beta}}=\abs{\Vec{C_2} . \Vec{X_{\beta}} - \Vec{X}}, 
            \Vec{D_{\delta}}=\abs{\Vec{C_3} . \Vec{X_{\gamma}} - \Vec{X}}
        \end{equation}
    
        \begin{equation}
            \Vec{X_1} = \Vec{X_{\alpha}} - \Vec{A_1} . (\Vec{D_{}\alpha}),
            \Vec{X_2} = \Vec{X_{\beta}} - \Vec{A_2} . (\Vec{D_{\beta}}),
            \Vec{X_3} = \Vec{X_{\delta}} - \Vec{A_3} . (\Vec{D_{\delta}})
        \end{equation}

        \begin{equation}
            \Vec{X}(t+1)=\frac{\Vec{X_1} + \Vec{X_2} + \Vec{X_3}}{3}
        \end{equation}

   After the hunting phase, the grey wolves start attacking the prey. To briefly explain the process of how GWO works, we first start by creating a random population of grey wolves/ individuals. Over many iterations, the individuals based on their hierarchy start to approximate the prey’s expected location. The individuals in the population interact with each other and shared information regarding the prey location till the optimal solution is found.
    \end{itemize}

    \subsection{Sperm Whale Algorithm}
    
    This Sperm Whale Algorithm~\cite{ebrahimi2016sperm} as the name suggests, makes use of how sperm whales interact in nature. This algorithm is based on how sperm whales use their incredibly good hearing to spot food and enemies, and how they use sound to communicate with one another. The main features of this algorithm can be stated as follows:
    
    \begin{itemize}
        \item Sperm whales come across two opposite poles in their cycle of breathing and feeding – the land for breathing and the sea for food. This feature has been used in the algorithm as two poles of answers, the Best and the Worst answers.
        \item The sperm whales travel in groups of 6-9 with males and females in the same group. When it’s the mating season, the males fight among themselves and the strongest male gets to mate with the females in the group. This has been used in the algorithm to choose better children for the next generations. Obviously, the candidates with higher Objective score wins the fight, and thus, the solution converges towards the optimal value with every generation.
    \end{itemize}
    
     The best and worst individuals in the population of the whales in each gathering are the $X_{best}$ and the $X_{worst}$,individually, showing the following relationships:

    \begin{equation}
        X_{center} = X_{worst} + c \times X_{best}
    \end{equation}
    \begin{equation}
        X_{reflex} =2X_{center} - X_{worst}    
    \end{equation}

    $X_{reflex}$  is situated outside the inquiry space, $c$ should diminish thus: 
    \begin{equation}
        c = r \times c_i
    \end{equation}

    where $c_i$ is the beginning focus factor and $r$ is the constriction coefficient that is less than $1$.

    \section{Applications of SI Techniques}

    The applications of Swarm Intelligence are limitless to be honest. It has put its name in various fields of research and other real-world usage. One of the biggest advancements in recent years has obviously been machine learning or deep learning and SI techniques have been extensively used to achieve greater results in the field. SI techniques have also shown development in Networking improving the cost of Wireless Sensor Networks (WSNs)~\cite{singh2021multi} and also topological issues, energy issues, energy issues, connectivity and coverage issues and localization issues. SI techniques have been very useful in speech processing~\cite{wright2013optimization} as well and have found  significant  reduction in background noise in audios for recordings. This is important because audio clips with  background noises and echo is unintelligent speech and SI techniques have improved a lot in this field. Techniques like PSO, ABC were seen to give better signal to noise ratio.

    Techniques like ACO are useful in image processing specifically in feature extraction from images~\cite{jino2019nature}. It has been seen that these techniques give a better perceptual graph while doing feature extraction from images. These techniques also have proved detrimental in image  segmentation and also been very useful in bioinformatics~\cite{handl2007multiobjective} by the use of automatic cluster detection and has been extensively used in computer vision for mammography in cancer risk  detection and breast cancer detection. Swarm intelligence has also been very useful in data mining in healthcare for better results in finding previous cases of diseases such as cancer, heart diseases, tumors and other health related problems. Swarm Intelligence has also been useful in logistics and transport, in efficient routing of cargo from different one destination to another, especially ACO, with its pheromones have very efficiently given desired outputs from the source to destination nodes. Similarly, for telecommunication as well, ACO has optimized the routing of different telecom networks and the efficient management of users in different networks has been handled. Beehive Algorithm are also useful in segmenting tasks for basic factory operations like transport from one portion to another, packaging etc. Different SI techniques have been useful for mass recruiting for any company because they use pheromones to attract and automatically cluster in more advantageous spots. 

    Furthermore, SI techniques have been used for automated machine learning by creating Deep Neural Networks (DNNs)~\cite{darwish2020survey}, which usually require severe expertise to be designed manually, social network analysis for detecting communities~\cite{biswas2016community},\cite{biswas2014community},\cite{biswas2015empirical}, \cite{garg2016evolutionary}. They have also shown better results in Resource Allocation; better processing of resources and managing assets in a strategic way. Swarm Intelligence also probably has countless other feats in several other fields like structural engineering~\cite{parpinelli2012comparison}. To describe them all would be a very lengthy task to continue, which we will discuss in later chapters, in more detail. To summarize in one sentence, Swarm Intelligence has definitely found its way to every field of study in this world.

    \section{\Large{Discussions and Conclusion}}
    
    In this chapter, we have presented the various population based swarm intelligence techniques that have been developed so far including particle swarm intelligence, firefly algorithm, bacteria colony algorithm, crow search algorithm, grey wolf optimization, sperm whale algorithm and their fundamental structures along with their mathematical models. SI based search technique and working of a few of these algorithms are highlighted. A brief explanation of applications of swarm intelligence techniques is done. Some major deductions from this chapter are listed below:
    \begin{enumerate}
    
        \item The major component that differentiates SI optimization algorithms from traditional optimization algorithms is the stochastic nature. Adaptive nature~\cite{biswas2015swarm} of the algorithms provide a tremendous potential to solve large scale, multi dimensional, complex problems more efficiently.
        \item Exploration and exploitation are the fundamental operations that in a SI search. The success rate of finding optimal solutions depend on the balance between exploration and exploitation. 
        \item As the nature of the optimization problem is different with one another, we need to choose which SI techniques to use based on various factors like working mechanism of the SI techniques, its parameters tuning capabilities, setting of parameters etc.
        \item While one or some SI algorithms may perform well on some optimization problems, it might not perform well on other optimization problems. So, according to our requirements we may need to modify parts of SI techniques, hybridize with other algorithms so that we can get the desired solution.
        \item Prior analysis of usage of SI techniques is crucial for applications. Though, statistical measures are available, but mostly lacks direct comparison~\cite{biswas2017regression}. Alternative techniques like visual analysis, which considers direct comparison of solutions would be useful~\cite{biswas2014visual,biswas2017analyzing}. 
    \end{enumerate}
    
    As  different swarm intelligence based algorithms provide a way to overcome different optimization problems efficiently and effectively as compared to traditional techniques, Swarm Intelligence techniques are becoming more  popular to solve various complex optimization problems. It has several advantages over traditional algorithms in terms having stochastic nature, efficient search of the search space using exploration and exploitation and faster convergence.

 \section*{Acknowledgements}   
    This work is supported by the Science and Engineering Board (SERB), Department of Science and Technology (DST) of the Government of India under Grant No. EEQ/2019/000657.
\nocite{*}
\bibliographystyle{unsrt}
\bibliography{references.bib}

\begin{thebibliography}{10}

\bibitem{blum2003metaheuristics}
Christian Blum and Andrea Roli.
\newblock Metaheuristics in combinatorial optimization: Overview and conceptual
  comparison.
\newblock {\em ACM computing surveys (CSUR)}, 35(3):268--308, 2003.

\bibitem{kennedy1995particle}
James Kennedy and Russell Eberhart.
\newblock Particle swarm optimization.
\newblock In {\em Proceedings of ICNN'95-international conference on neural
  networks}, volume~4, pages 1942--1948. IEEE, 1995.

\bibitem{yang2010firefly}
Xin-She Yang.
\newblock Firefly algorithm, levy flights and global optimization.
\newblock In {\em Research and development in intelligent systems XXVI}, pages
  209--218. Springer, 2010.

\bibitem{binitha2012survey}
S~Binitha, S~Siva Sathya, et~al.
\newblock A survey of bio inspired optimization algorithms.
\newblock {\em International journal of soft computing and engineering},
  2(2):137--151, 2012.

\bibitem{biswas2013physics}
Anupam Biswas, KK~Mishra, Shailesh Tiwari, and AK~Misra.
\newblock Physics-inspired optimization algorithms: a survey.
\newblock {\em Journal of Optimization}, 2013, 2013.

\bibitem{houssein2019nature}
Essam~H Houssein, Mina Younan, and Aboul~Ella Hassanien.
\newblock Nature-inspired algorithms: A comprehensive review.
\newblock {\em Hybrid Computational Intelligence}, pages 1--25, 2019.

\bibitem{zhiheng2021flamingo}
Wang Zhiheng and Liu Jianhua.
\newblock Flamingo search algorithm: A new swarm intelligence optimization
  algorithm.
\newblock {\em IEEE Access}, 9(1):88564--88582, 2021.

\bibitem{miarnaeimi2021horse}
Farid MiarNaeimi, Gholamreza Azizyan, and Mohsen Rashki.
\newblock Horse herd optimization algorithm: A nature-inspired algorithm for
  high-dimensional optimization problems.
\newblock {\em Knowledge-Based Systems}, 213:106711, 2021.

\bibitem{khishe2020chimp}
M~Khishe and Mohammad~Reza Mosavi.
\newblock Chimp optimization algorithm.
\newblock {\em Expert systems with applications}, 149:113338, 2020.

\bibitem{hayyolalam2020black}
Vahideh Hayyolalam and Ali Asghar~Pourhaji Kazem.
\newblock Black widow optimization algorithm: a novel meta-heuristic approach
  for solving engineering optimization problems.
\newblock {\em Engineering Applications of Artificial Intelligence}, 87:103249,
  2020.

\bibitem{xue2020novel}
Jiankai Xue and Bo~Shen.
\newblock A novel swarm intelligence optimization approach: sparrow search
  algorithm.
\newblock {\em Systems Science $\&$ Control Engineering}, 8(1):22--34, 2020.

\bibitem{dhiman2020novel}
Gaurav Dhiman, Meenakshi Garg, Atulya Nagar, Vijay Kumar, and Mohammad
  Dehghani.
\newblock A novel algorithm for global optimization: Rat swarm optimizer.
\newblock {\em Journal of Ambient Intelligence and Humanized Computing}, pages
  1--26, 2020.

\bibitem{shadravan2019sailfish}
S~Shadravan, HR~Naji, and Vahid~Khatibi Bardsiri.
\newblock The sailfish optimizer: A novel nature-inspired metaheuristic
  algorithm for solving constrained engineering optimization problems.
\newblock {\em Engineering Applications of Artificial Intelligence}, 80:20--34,
  2019.

\bibitem{al2018meerkat}
Ahmed T~Sadiq Al-Obaidi, Hasanen~S Abdullah, and Zied~O Ahmed.
\newblock Meerkat clan algorithm: A new swarm intelligence algorithm.
\newblock {\em Indonesian Journal of Electrical Engineering and Computer
  Science}, 10(1):354--360, 2018.

\bibitem{mirjalili2018grasshopper}
Seyedeh~Zahra Mirjalili, Seyedali Mirjalili, Shahrzad Saremi, Hossam Faris, and
  Ibrahim Aljarah.
\newblock Grasshopper optimization algorithm for multi-objective optimization
  problems.
\newblock {\em Applied Intelligence}, 48(4):805--820, 2018.

\bibitem{mirjalili2017salp}
Seyedali Mirjalili, Amir~H Gandomi, Seyedeh~Zahra Mirjalili, Shahrzad Saremi,
  Hossam Faris, and Seyed~Mohammad Mirjalili.
\newblock Salp swarm algorithm: A bio-inspired optimizer for engineering design
  problems.
\newblock {\em Advances in Engineering Software}, 114:163--191, 2017.

\bibitem{al2017camel}
Ahmed T~Sadiq Al-Obaidi, Hasanen~S Abdullah, et~al.
\newblock Camel herds algorithm: A new swarm intelligent algorithm to solve
  optimization problems.
\newblock {\em International Journal on Perceptive and Cognitive Computing},
  3(1), 2017.

\bibitem{wang2017duck}
Weihong Wang, Sentang Wu, Ke~Lu, et~al.
\newblock Duck pack algorithm—a new swarm intelligence algorithm for route
  planning based on imprinting behavior.
\newblock In {\em 2017 29th Chinese Control And Decision Conference (CCDC)},
  pages 2392--2396. IEEE, 2017.

\bibitem{mirjalili2016dragonfly}
Seyedali Mirjalili.
\newblock Dragonfly algorithm: a new meta-heuristic optimization technique for
  solving single-objective, discrete, and multi-objective problems.
\newblock {\em Neural Computing and Applications}, 27(4):1053--1073, 2016.

\bibitem{ebrahimi2016sperm}
A~Ebrahimi and E~Khamehchi.
\newblock Sperm whale algorithm: an effective metaheuristic algorithm for
  production optimization problems.
\newblock {\em Journal of Natural Gas Science and Engineering}, 29:211--222,
  2016.

\bibitem{wu2016dolphin}
Tian-qi Wu, Min Yao, and Jian-hua Yang.
\newblock Dolphin swarm algorithm.
\newblock {\em Frontiers of Information Technology $\&$ Electronic
  Engineering}, 17(8):717--729, 2016.

\bibitem{askarzadeh2016novel}
Alireza Askarzadeh.
\newblock A novel metaheuristic method for solving constrained engineering
  optimization problems: crow search algorithm.
\newblock {\em Computers $\&$ Structures}, 169:1--12, 2016.

\bibitem{mirjalili2015ant}
Seyedali Mirjalili.
\newblock The ant lion optimizer.
\newblock {\em Advances in engineering software}, 83:80--98, 2015.

\bibitem{wang2015elephant}
Gai-Ge Wang, Suash Deb, and Leandro dos~S Coelho.
\newblock Elephant herding optimization.
\newblock In {\em 2015 3rd International Symposium on Computational and
  Business Intelligence (ISCBI)}, pages 1--5. IEEE, 2015.

\bibitem{mirjalili2015moth}
Seyedali Mirjalili.
\newblock Moth-flame optimization algorithm: A novel nature-inspired heuristic
  paradigm.
\newblock {\em Knowledge-based systems}, 89:228--249, 2015.

\bibitem{mirjalili2014grey}
Seyedali Mirjalili, Seyed~Mohammad Mirjalili, and Andrew Lewis.
\newblock Grey wolf optimizer.
\newblock {\em Advances in engineering software}, 69:46--61, 2014.

\bibitem{goel2014pigeon}
Shruti Goel.
\newblock Pigeon optimization algorithm: A novel approach for solving
  optimization problems.
\newblock In {\em 2014 International Conference on Data Mining and Intelligent
  Computing (ICDMIC)}, pages 1--5. IEEE, 2014.

\bibitem{bansal2014spider}
Jagdish~Chand Bansal, Harish Sharma, Shimpi~Singh Jadon, and Maurice Clerc.
\newblock Spider monkey optimization algorithm for numerical optimization.
\newblock {\em Memetic computing}, 6(1):31--47, 2014.

\bibitem{cuevas2013swarm}
Erik Cuevas, Miguel Cienfuegos, Daniel Zald{\'\i}var, and Marco
  P{\'e}rez-Cisneros.
\newblock A swarm optimization algorithm inspired in the behavior of the
  social-spider.
\newblock {\em Expert Systems with Applications}, 40(16):6374--6384, 2013.

\bibitem{niu2012bacterial}
Ben Niu and Hong Wang.
\newblock Bacterial colony optimization.
\newblock {\em Discrete Dynamics in Nature and Society}, 2012, 2012.

\bibitem{nguyen2012zombie}
Hoang~Thanh Nguyen and Bir Bhanu.
\newblock Zombie survival optimization: A swarm intelligence algorithm inspired
  by zombie foraging.
\newblock In {\em Proceedings of the 21st International Conference on Pattern
  Recognition (ICPR2012)}, pages 987--990. IEEE, 2012.

\bibitem{yang2010new}
Xin-She Yang.
\newblock A new metaheuristic bat-inspired algorithm.
\newblock In {\em Nature inspired cooperative strategies for optimization
  (NICSO 2010)}, pages 65--74. Springer, 2010.

\bibitem{tan2010fireworks}
Ying Tan and Yuanchun Zhu.
\newblock Fireworks algorithm for optimization.
\newblock In {\em International conference in swarm intelligence}, pages
  355--364. Springer, 2010.

\bibitem{yang2009cuckoo}
Xin-She Yang and Suash Deb.
\newblock Cuckoo search via l{\'e}vy flights.
\newblock In {\em 2009 World congress on nature $\&$ biologically inspired
  computing (NaBIC)}, pages 210--214. Ieee, 2009.

\bibitem{rashedi2009gsa}
Esmat Rashedi, Hossein Nezamabadi-Pour, and Saeid Saryazdi.
\newblock Gsa: a gravitational search algorithm.
\newblock {\em Information sciences}, 179(13):2232--2248, 2009.

\bibitem{krishnanand2009glowworm}
KN~Krishnanand and Debasish Ghose.
\newblock Glowworm swarm optimisation: a new method for optimising multi-modal
  functions.
\newblock {\em International Journal of Computational Intelligence Studies},
  1(1):93--119, 2009.

\bibitem{chu2008fast}
Ying Chu, Hua Mi, Huilian Liao, Zhen Ji, and QH~Wu.
\newblock A fast bacterial swarming algorithm for high-dimensional function
  optimization.
\newblock In {\em 2008 IEEE congress on evolutionary computation (ieee world
  congress on computational intelligence)}, pages 3135--3140. IEEE, 2008.

\bibitem{chu2006cat}
Shu-Chuan Chu, Pei-Wei Tsai, and Jeng-Shyang Pan.
\newblock Cat swarm optimization.
\newblock In {\em Pacific Rim international conference on artificial
  intelligence}, pages 854--858. Springer, 2006.

\bibitem{hancer2012artificial}
Emrah Hancer, Celal Ozturk, and Dervis Karaboga.
\newblock Artificial bee colony based image clustering method.
\newblock In {\em 2012 IEEE congress on evolutionary computation}, pages 1--5.
  IEEE, 2012.

\bibitem{wedde2004beehive}
Horst~F Wedde, Muddassar Farooq, and Yue Zhang.
\newblock Beehive: An efficient fault-tolerant routing algorithm inspired by
  honey bee behavior.
\newblock In {\em International Workshop on Ant Colony Optimization and Swarm
  Intelligence}, pages 83--94. Springer, 2004.

\bibitem{dorigo2006ant}
Marco Dorigo, Mauro Birattari, and Thomas Stutzle.
\newblock Ant colony optimization.
\newblock {\em IEEE computational intelligence magazine}, 1(4):28--39, 2006.

\bibitem{sait2013cell}
Sadiq~M Sait, Ahmad~T Sheikh, and Aiman~H El-Maleh.
\newblock Cell assignment in hybrid cmos/nanodevices architecture using a
  pso/sa hybrid algorithm.
\newblock {\em Journal of applied research and technology}, 11(5):653--664,
  2013.

\bibitem{kuo2013integration}
RJ~Kuo and CW~Hong.
\newblock Integration of genetic algorithm and particle swarm optimization for
  investment portfolio optimization.
\newblock {\em Applied mathematics $\&$ information sciences}, 7(6):2397, 2013.

\bibitem{chen2011solving}
Shyi-Ming Chen and Chih-Yao Chien.
\newblock Solving the traveling salesman problem based on the genetic simulated
  annealing ant colony system with particle swarm optimization techniques.
\newblock {\em Expert Systems with Applications}, 38(12):14439--14450, 2011.

\bibitem{jau2013modified}
You-Min Jau, Kuo-Lan Su, Chia-Ju Wu, and Jin-Tsong Jeng.
\newblock Modified quantum-behaved particle swarm optimization for parameters
  estimation of generalized nonlinear multi-regressions model based on choquet
  integral with outliers.
\newblock {\em Applied Mathematics and Computation}, 221:282--295, 2013.

\bibitem{chen2012discrete}
Min Chen and Simone~A Ludwig.
\newblock Discrete particle swarm optimization with local search strategy for
  rule classification.
\newblock In {\em 2012 Fourth World Congress on Nature and Biologically
  Inspired Computing (NaBIC)}, pages 162--167. IEEE, 2012.

\bibitem{biswas2018particle}
Anupam Biswas, Bhaskar Biswas, Anoj Kumar, and KK~Mishra.
\newblock Particle swarm optimisation with time varying cognitive avoidance
  component.
\newblock {\em International Journal of Computational Science and Engineering},
  16(1):27--41, 2018.

\bibitem{biswas2013particle}
Anupam Biswas, Anoj Kumar, and KK~Mishra.
\newblock Particle swarm optimization with cognitive avoidance component.
\newblock In {\em 2013 International Conference on Advances in Computing,
  Communications and Informatics (ICACCI)}, pages 149--154. IEEE, 2013.

\bibitem{qiu2013novel}
Chenye Qiu, Chunlu Wang, and Xingquan Zuo.
\newblock A novel multi-objective particle swarm optimization with k-means
  based global best selection strategy.
\newblock {\em International Journal of Computational Intelligence Systems},
  6(5):822--835, 2013.

\bibitem{biswas2013improved}
Anupam Biswas, AV~Lakra, Sharad Kumar, and Avjeet Singh.
\newblock An improved random inertia weighted particle swarm optimization.
\newblock In {\em 2013 International Symposium on Computational and Business
  Intelligence}, pages 96--99. IEEE, 2013.

\bibitem{chuang2011chaotic}
Li-Yeh Chuang, Sheng-Wei Tsai, and Cheng-Hong Yang.
\newblock Chaotic catfish particle swarm optimization for solving global
  numerical optimization problems.
\newblock {\em Applied mathematics and computation}, 217(16):6900--6916, 2011.

\bibitem{fister2013comprehensive}
Iztok Fister, Iztok Fister~Jr, Xin-She Yang, and Janez Brest.
\newblock A comprehensive review of firefly algorithms.
\newblock {\em Swarm and Evolutionary Computation}, 13:34--46, 2013.

\bibitem{singh2021multi}
Omkar Singh, Vinay Rishiwal, Rashmi Chaudhry, and Mano Yadav.
\newblock Multi-objective optimization in wsn: Opportunities and challenges.
\newblock {\em Wireless Personal Communications}, 121(1):127--152, 2021.

\bibitem{wright2013optimization}
Stephen~J Wright, Dimitri Kanevsky, Li~Deng, Xiaodong He, Georg Heigold, and
  Haizhou Li.
\newblock Optimization algorithms and applications for speech and language
  processing.
\newblock {\em IEEE Transactions on Audio, Speech, and Language Processing},
  21(11):2231--2243, 2013.

\bibitem{jino2019nature}
SR~Jino~Ramson, K~Lova~Raju, S~Vishnu, and Theodoros Anagnostopoulos.
\newblock Nature inspired optimization techniques for image processing—a
  short review.
\newblock {\em Nature Inspired Optimization Techniques for Image Processing
  Applications}, pages 113--145, 2019.

\bibitem{handl2007multiobjective}
Julia Handl, Douglas~B Kell, and Joshua Knowles.
\newblock Multiobjective optimization in bioinformatics and computational
  biology.
\newblock {\em IEEE/ACM Transactions on computational biology and
  bioinformatics}, 4(2):279--292, 2007.

\bibitem{darwish2020survey}
Ashraf Darwish, Aboul~Ella Hassanien, and Swagatam Das.
\newblock A survey of swarm and evolutionary computing approaches for deep
  learning.
\newblock {\em Artificial Intelligence Review}, 53(3):1767--1812, 2020.

\bibitem{biswas2016community}
Anupam Biswas.
\newblock {\em Community detection in social networks using agglomerative and
  evalutionary techniques}.
\newblock PhD thesis, 2016.

\bibitem{biswas2014community}
A~Biswas, P~Gupta, M~Modi, and B~Biswas.
\newblock Community detection in multiple featured social network using swarm
  intelligence.
\newblock In {\em International Conference on Communication and Computing (ICC
  2014), Bangalore}, 2014.

\bibitem{biswas2015empirical}
Anupam Biswas, Pawan Gupta, Mradul Modi, and Bhaskar Biswas.
\newblock An empirical study of some particle swarm optimizer variants for
  community detection.
\newblock In {\em Advances in Intelligent Informatics}, pages 511--520.
  Springer, 2015.

\bibitem{garg2016evolutionary}
Abhishek Garg, Anupam Biswas, and Bhaskar Biswas.
\newblock Evolutionary computation techniques for community detection in social
  network analysis.
\newblock In {\em Advanced Methods for Complex Network Analysis}, pages
  266--284. IGI Global, 2016.

\bibitem{parpinelli2012comparison}
Rafael~S Parpinelli, F{\'a}bio~R Teodoro, and Heitor~S Lopes.
\newblock A comparison of swarm intelligence algorithms for structural
  engineering optimization.
\newblock {\em International Journal for Numerical Methods in Engineering},
  91(6):666--684, 2012.

\bibitem{biswas2015swarm}
Anupam Biswas and Bhaskar Biswas.
\newblock Swarm intelligence techniques and their adaptive nature with
  applications.
\newblock In {\em Complex System Modelling and Control Through Intelligent Soft
  Computations}, pages 253--273. Springer, 2015.

\bibitem{biswas2017regression}
Anupam Biswas and Bhaskar Biswas.
\newblock Regression line shifting mechanism for analyzing evolutionary
  optimization algorithms.
\newblock {\em Soft Computing}, 21(21):6237--6252, 2017.

\bibitem{biswas2014visual}
Anupam Biswas and Bhaskar Biswas.
\newblock Visual analysis of evolutionary optimization algorithms.
\newblock In {\em 2014 2nd International Symposium on Computational and
  Business Intelligence}, pages 81--84. IEEE, 2014.

\bibitem{biswas2017analyzing}
Anupam Biswas and Bhaskar Biswas.
\newblock Analyzing evolutionary optimization and community detection
  algorithms using regression line dominance.
\newblock {\em Information Sciences}, 396:185--201, 2017.

\bibitem{8869366}
J.~Revathi, V.P Eswaramurthy, and P.~Padmavathi.
\newblock Bacterial colony optimization for data clustering.
\newblock In {\em 2019 IEEE International Conference on Electrical, Computer
  and Communication Technologies (ICECCT)}, pages 1--4, 2019.

\bibitem{10.1007/978-3-642-31837-5_73}
Ben Niu and Hong" Wang.
\newblock Bacterial colony optimization: Principles and foundations.
\newblock In {\em Emerging Intelligent Computing Technology and Applications},
  pages 501--506, 2012.

\bibitem{ASKARZADEH20161}
Alireza Askarzadeh.
\newblock A novel metaheuristic method for solving constrained engineering
  optimization problems: Crow search algorithm.
\newblock {\em Computers $\&$ Structures}, 169:1--12, 2016.

\bibitem{9195808}
Abdelazim~G. Hussien, Mohamed Amin, Mingjing Wang, Guoxi Liang, Ahmed Alsanad,
  Abdu Gumaei, and Huiling Chen.
\newblock Crow search algorithm: Theory, recent advances, and applications.
\newblock {\em IEEE Access}, 8:173548--173565, 2020.

\bibitem{Zolghadr-Asli2018}
Babak Zolghadr-Asli, Omid Bozorg-Haddad, and Xuefeng Chu.
\newblock {\em Crow Search Algorithm (CSA)}, pages 143--149.
\newblock Springer Singapore, Singapore, 2018.

\bibitem{NIU201937}
Peifeng Niu, Songpeng Niu, Nan liu, and Lingfang Chang.
\newblock The defect of the grey wolf optimization algorithm and its
  verification method.
\newblock {\em Knowledge-Based Systems}, 171:37--43, 2019.

\bibitem{MIRJALILI201446}
Seyedali Mirjalili, Seyed~Mohammad Mirjalili, and Andrew Lewis.
\newblock Grey wolf optimizer.
\newblock {\em Advances in Engineering Software}, 69:46--61, 2014.

\bibitem{article}
Seyedali Mirjalili, Seyed Mirjalili, and Andrew Lewis.
\newblock Grey wolf optimizer.
\newblock {\em Advances in Engineering Software}, 69:46–61, 03 2014.

\bibitem{EBRAHIMI2016211}
A.~Ebrahimi and E.~Khamehchi.
\newblock Sperm whale algorithm: An effective metaheuristic algorithm for
  production optimization problems.
\newblock {\em Journal of Natural Gas Science and Engineering}, 29:211--222,
  2016.

\bibitem{10.1007/978-3-030-53956-6_1}
Jian Yang, Liang Qu, Yang Shen, Yuhui Shi, Shi Cheng, Junfeng Zhao, and
  Xiaolong Shen.
\newblock Swarm intelligence in data science: Applications, opportunities and
  challenges.
\newblock In Ying Tan, Yuhui Shi, and Milan Tuba, editors, {\em Advances in
  Swarm Intelligence}, pages 3--14, Cham, 2020. Springer International
  Publishing.

\bibitem{applcationSwarm2013}
Yudong Zhang, Praveen Agarwal, Vishal Bhatnagar, Saeed Balochian, and Jie Yan.
\newblock Swarm intelligence and its applications, 2013.

\bibitem{applicationSwarm2013}
R.~Ganesan M.~Vergin Raja~Sarobin.
\newblock Swarm intelligence in wireless sensor networks: A survey.
\newblock {\em International Journal of Pure and Applied Mathematics}, 101,
  2015.

\bibitem{7380645}
Khumukcham~Usharani Devi, Dipjyoti Sarma, and Romesh Laishram.
\newblock Swarm intelligence based computing techniques in speech enhancement.
\newblock In {\em 2015 International Conference on Green Computing and Internet
  of Things (ICGCIoT)}, pages 1199--1203, 2015.

\bibitem{articleSwarm}
Xiaodong Zhuang and Nikos Mastorakis.
\newblock Image processing with the artificial swarm intelligence.
\newblock {\em WSEAS Transactions on Computers}, 4:333--341, 04 2005.

\bibitem{inbook}
Santosh Kumar, Deepanwita Datta, and Sanjay Singh.
\newblock {\em Swarm Intelligence for Biometric Feature Optimization}, pages
  147--181.
\newblock 01 2015.

\bibitem{articleManet}
Ihtiram Raza~Khan Mehtab~Alam, Asif Hameed~Khan.
\newblock Swarm intelligence in manets: A survey.
\newblock {\em International Journal of Emerging Research in Management $\&$
  Technology}, 5:141--150, 05 2016.

\bibitem{inproceedings}
Nandini Nayar, Sachin Ahuja, and Dr.~Shaily Jain.
\newblock Swarm intelligence and data mining: a review of literature and
  applications in healthcare.
\newblock pages 1--7, 06 2019.

\bibitem{inproceedingSwarm}
Davide Anghinolfi, Antonio Boccalatte, Alberto Grosso, Massimo Paolucci, Andrea
  Passadore, and Christian Vecchiola.
\newblock A swarm intelligence method applied to manufacturing scheduling.
\newblock pages 65--70, 01 2007.

\end{thebibliography}
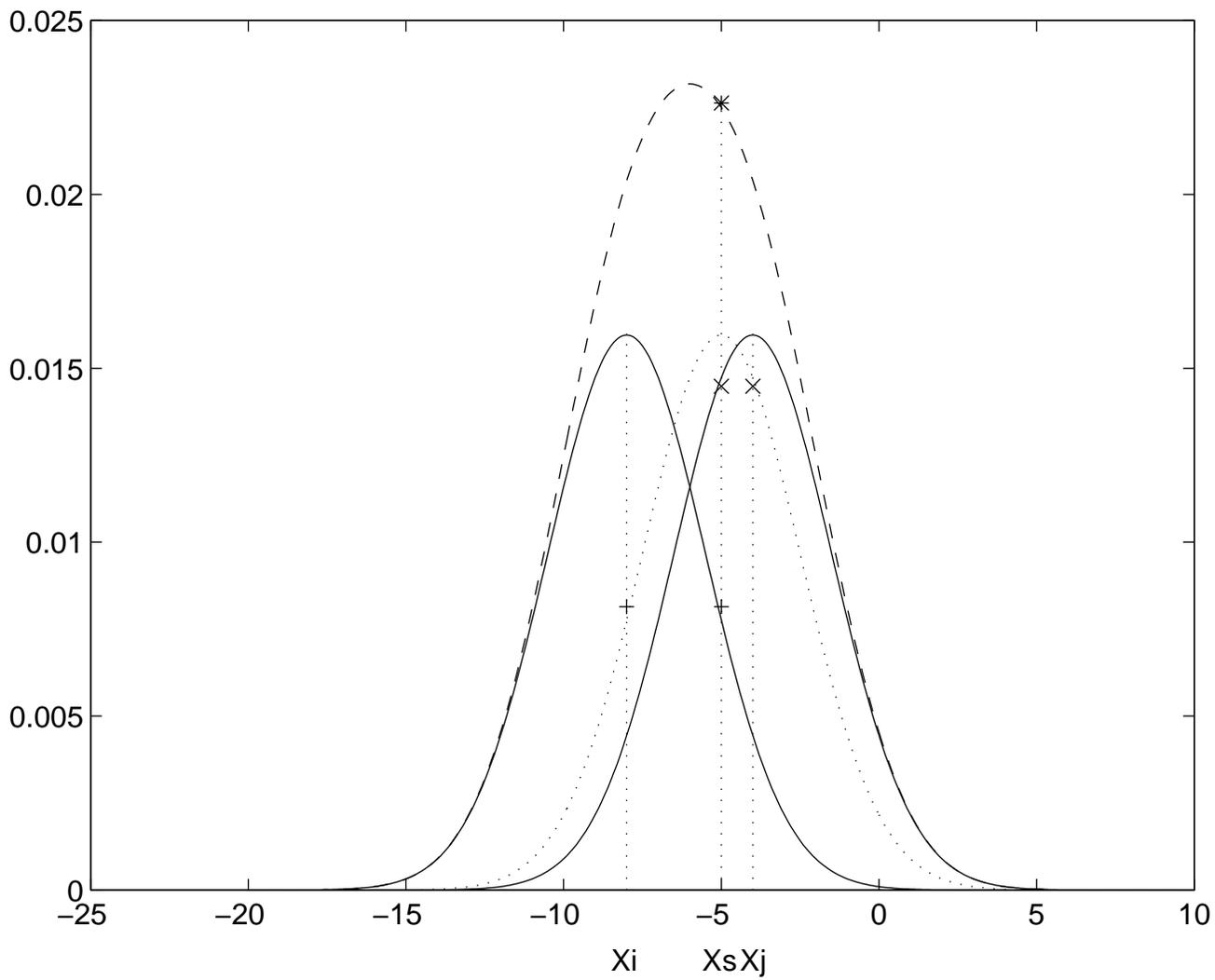
\end{document}